\documentclass[letterpaper, 10 pt, conference]{ieeeconf}  % Comment this line out if you need a4paper
\usepackage{url}
\usepackage{graphicx}

\usepackage{amsmath,amsfonts}
\usepackage{algorithmic}
\usepackage{algorithm}

\usepackage{array}
\usepackage[caption=false,font=normalsize,labelfont=sf,textfont=sf]{subfig}
\usepackage{textcomp}
\usepackage{stfloats}
\usepackage{hyperref}
\usepackage{verbatim}
\usepackage{cite}
\usepackage{amssymb}
\usepackage{color}
\usepackage{cleveref}
\newtheorem{lemma}{Lemma}

\IEEEoverridecommandlockouts                              % This command is only needed if 
\title{\LARGE \bf
Robust Self-Reconfiguration for Fault-Tolerant Control of Modular Aerial Robot Systems
}

\author{Rui Huang, Siyu Tang, Zhiqian Cai,  Lin Zhao% <-this % stops a space
\thanks{Rui Huang,  Siyu Tang, and Lin Zhao are with the Department of Electrical and Computer Engineering, National University of Singapore, Singapore 117583 (Email: 
        {ruihuang@u.nus.edu, e1352616@u.nus.edu, elezhli@nus.edu.sg}).
        Zhiqian Cai is with the Engineering Design and Innovation Centre, National University of Singapore, Singapore 117583 (Email: 
        {e1391152@u.nus.edu}). This work was supported by the Singapore Ministry of Education Tier 2 AcRF under A-8001889-00-00. (Corresponding author: Lin Zhao)
}
}
\begin{document}

\maketitle
\thispagestyle{empty}
\pagestyle{empty}

%%%%%%%%%%%%%%%%%%%%%%%%%%%%%%%%%%%%%%%%%%%%%%%%%%%%%%%%%%%%%%%%%%%%%%%%%%%%%%%%
\begin{abstract}

Modular Aerial Robotic Systems (MARS) consist of multiple drone units assembled into a single, integrated rigid flying platform. With inherent redundancy, MARS can self-reconfigure into different configurations to mitigate rotor or unit failures and maintain stable flight. However, existing works on MARS self-reconfiguration often overlook the practical controllability of intermediate structures formed during the reassembly process, which limits their applicability. In this paper, we address this gap by considering the control-constrained dynamic model of MARS and proposing a robust and efficient self-reconstruction algorithm that maximizes the controllability margin at each intermediate stage. Specifically, we develop algorithms to compute optimal, controllable disassembly and assembly sequences, enabling robust self-reconfiguration. Finally, we validate our method in several challenging fault-tolerant self-reconfiguration scenarios, demonstrating significant improvements in both controllability and trajectory tracking while reducing the number of assembly steps. The videos and source code of this work are available at \url{https://github.com/RuiHuangNUS/MARS-Reconfig/}

\end{abstract}

%%%%%%%%%%%%%%%%%%%%%%%%%%%%%%%%%%%%%%%%%%%%%%%%%%%%%%%%%%%%%%%%%%%%%%%%%%%%%%%%

\section{Introduction}
\begin{figure}[!t]
\centering
\includegraphics[width=3.4in]{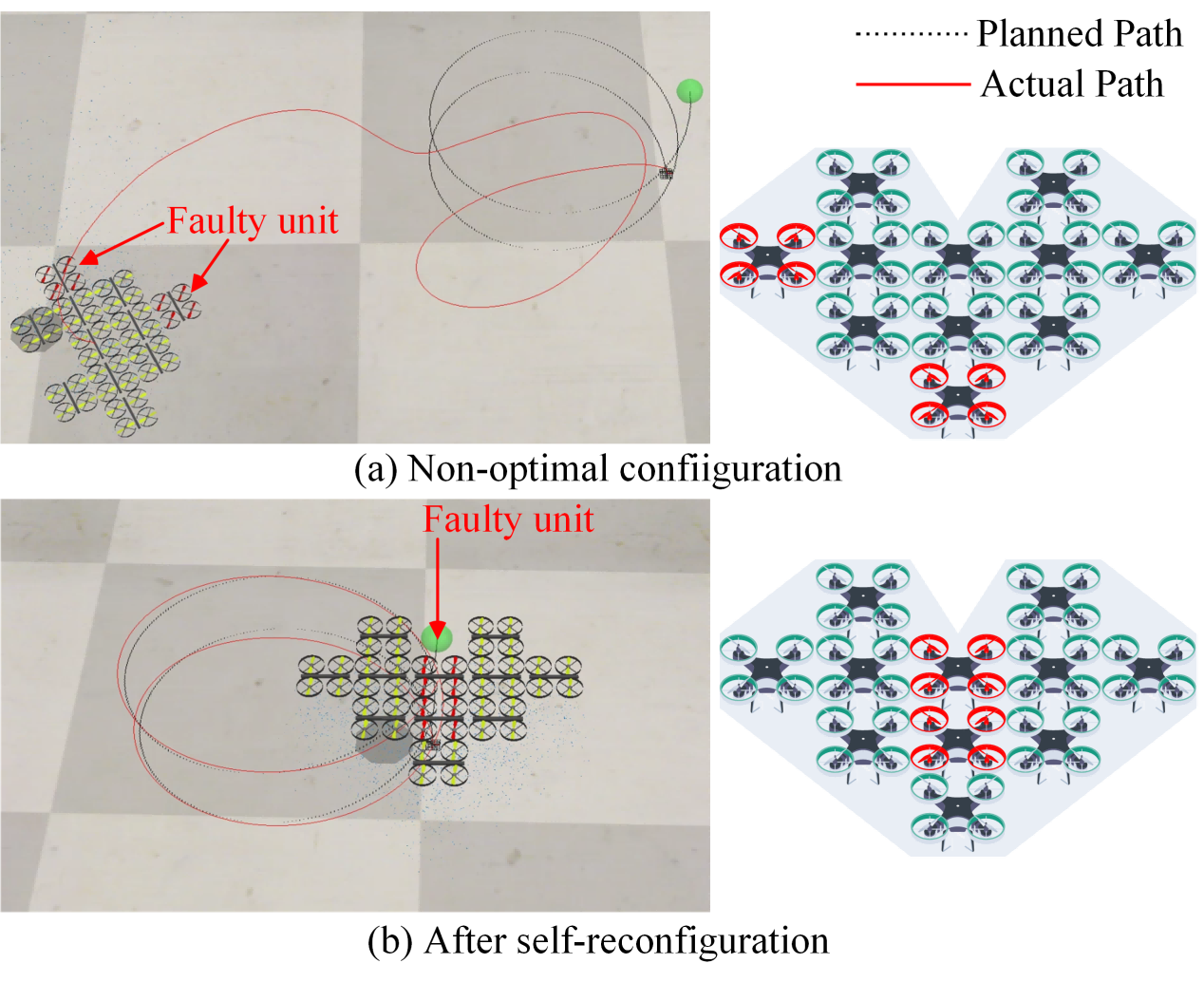}
\vspace{-3 mm}
\caption{MARS is tasked to track a spiral trajectory with two faulty units. The faulty propellers are marked red. (a) MARS crashed after the complete failure of two units (all rotors are broken). (b) MARS can track trajectories after self-reconfiguration.}
\label{fig:MARS}
\vspace{-6mm}
\end{figure}

Modular Aerial Robotic Systems (MARS) offer fast and flexible navigation and task execution by dynamically disassembling and reassembling according to environmental conditions, such as navigation through cluttered or narrow spaces~\cite{zhang2024design,li2019modquad,wang2024trust,saldana2018modquad,litman2021vision}. 
However, aerial robots are vulnerable to various faults, such as multiple rotor failures~\cite{9496133}, which can severely degrade overall control performance if not properly addressed. Fig.~\ref{fig:MARS}(a) illustrates a simulation where MARS fails to track a spiral trajectory after two units experience faults (drones with red propellers). With inherent redundancy, effective fault-tolerant control allows MARS to reconfigure its assembly structure, improving control performance. Fig.~\ref{fig:MARS}(b) demonstrates enhanced tracking performance after reconfiguration, where faulty drones are encompassed by functional ones to restore controllability. 

Achieving such fault-tolerant reconfiguration for various robotic swarms is a challenging problem that has attracted significant attention in recent research~\cite{shi2020aerial,9962372,10404022,huang2024adaptive}. Existing works on MARS self-reconfiguration focus on optimizing structure and reconfiguration strategies, but many do not adequately address the control and coordination required at each stage of the process, particularly when considering the controllability of intermediate configurations. Crucially, the intermediate stages of the reconfiguration process must maintain controllability for each separated and combined part of the MARS under realistic physical limits of control input. 

Recent work~\cite{gandhi2020self} proposed a self-reconfiguration technique for rotor failures, which aims to enable mission continuation while mitigating the effects of rotor failures. This method attempts to reduce the impact of failures on the trajectory by applying a mixed integer linear program (MILP) to determine an appropriate module placement. However, this optimization approach has certain limitations. The assembly tree of the failure sequence must be computed offline in advance to access the configuration tree in the event of a failure at a specific location, which limits real-time adaptability. Their method focuses on connecting a functional module to a unit experiencing a partial failure, such as when only one rotor in a quadcopter fails, but it does not address scenarios involving complete module failures or multiple drone unit failures. Additionally, the final assembly configuration could potentially be improved by adjusting the orientation of the partially failed modules to enhance control authority. Moreover, it does not fully account for uncontrollable factors, and its effectiveness is limited by the chosen objective function and the accuracy of the physical simulation engine. 

To address the safety and controllability challenges in the self-reconfiguration process while minimizing the number of reconfiguration steps, we propose a controllability theory-based self-reconfiguration algorithm that incorporates a more practical MARS model with constrained control inputs. Compared to existing state-of-the-art methods, our approach offers an interpretable and quantifiable measure to determine the robustness of configurations. Specifically, our key contributions include:
\begin{enumerate}
    \item We propose an efficient controllability analysis for MARS using a quasi-static model under control constraints. Previous methods~\cite{gandhi2020self} overlook control input limitations, leading to inaccurate characterization of controllability in practical systems. Our analysis leverages positive controllability theory~\cite{brammer1972controllability}  to calculate a single controllability margin, which quantifies the available control authority.
    \item  We propose a novel controllability margin (CM)--based method for calculating optimal self-reconfiguration sequences. This method identifies the optimal assembly configuration by maximizing CM and determines the smallest controllable assembly structure containing the faulty unit in terms of the number of drone units. Moreover, we developed algorithms to compute disassembly and assembly sequences that guarantee controllability throughout the entire reconfiguration process. Extensive experiments demonstrate that our approach can ensure the controllability of the intermediate sub-assemblies during the reconfiguration while the method in~\cite{gandhi2020self} cannot. Meanwhile, our approaches require significantly less assembly and disassembly to achieve the same target configuration. 
\end{enumerate}

\section{Preliminaries}

We briefly describe the dynamical model of MARS used in our analysis and algorithm design. We assume that all the assemblies and disassemblies during the reconfiguration occur in a \textit{quasi-static hovering state} (and in addition, the orientation of the global frame and body frame are always aligned), allowing us to simplify the dynamic model of MARS to the following linear system: 
\begin{equation}
\dot{\mathbf{x}}=\mathbf{A}\mathbf{x}+\mathbf{B}\left(\mathbf{u}_f-\mathbf{g}\right),
\label{eq:system}
\end{equation}
with
\begin{equation}
\begin{aligned}&\mathbf{x}=[p_{z_{e}}\quad\phi\quad\theta\quad\psi\quad\nu_{z_{e}}\quad\omega_{x_{b}}\quad\omega_{y_{b}}\quad\omega_{z_{b}}]^{\mathrm{T}}\in\mathbb{R}^{8},\\
&\mathbf{u}_{f}=[F\quad M_{x}\quad M_{y}\quad M_{z}]^{\mathrm{T}}\in\mathbb{R}^{4},\\
&\mathbf{g}=[nmg\quad0\quad0\quad0]^{\mathrm{T}}\in\mathbb{R}^{4},\\
&\mathbf{A}=\begin{bmatrix}\mathbf{0}_{4\times4}&\mathbf{I}_{4}\\\mathbf{0}_{4\times4}&\mathbf{0}_{4\times4}\end{bmatrix}\in\mathbb{R}^{8\times8},\\
&\mathbf{B}=\begin{bmatrix}\mathbf{0}_{4\times4} \\
\mathrm{diag}\left(-nm,\mathbf{J}\right)^{-1}\end{bmatrix}
\in\mathbb{R}^{8\times4},\\
&\mathbf{J}=\mathrm{diag}\left(J_{xx},J_{yy},J_{zz}\right)\in\mathbb{R}^{3\times3},
\end{aligned}
\label{eq:1}
\end{equation}
where the state vector $\mathbf{x}$ includes the MARS center of mass (CoM) global vertical position $p_{z_{e}}$ in the earth coordinate frame, roll angle $\phi$, pitch angle $\theta$, yaw angle $\psi$, and the corresponding linear velocity $\nu_{z_{e}}$. It also includes the angular velocities $\omega_{x_{b}}, \omega_{y_{b}}, \omega_{z_{b}}$ along the coordinate axes in the body coordinate frame. Moreover, $F$ represents the collective thrust applied to the MARS and $[M_x, M_y, M_z]$ denotes the collective torques. The parameters $n$, $m$, $g$, and $\mathbf{J}$ refer to the total number of modular drone units, the mass of a single drone unit, the gravitational acceleration constant, and the moment of inertia, respectively.

We consider the realistic control limits, where each rotor's thrust force satisfies $T_{i}\in[0,K_{i}]$, where $K_i>0$, $i=1,2\cdots n_{r}$, and $n_{r}$ denotes the total number of rotors in the MARS. Stacking each rotor's thrust into a vector, we  define the force vector $\mathbf{f}=\left[T_{1},T_{2}\cdots T_{n_{r}}\right]^{\mathrm{T}}$, which is constrained as follows:
\begin{equation}
    \hspace*{-0.20cm}
    \mathbf{f}\!\in\! U\!_f\!=\!\left\{\mathbf{f}\!=\!\left[T\!_1\!,\!T\!_2\!\cdots T\!_{n_r}\!\right]^\mathrm{T}\!\mid0\!\leq T\!_i\!\leq K\!_i,i\!=\!1\!,2\!\cdots \!n_r\!\right\}
\label{eq:f}
\end{equation}
Based on the geometric layout of each drone unit, the mapping relationship between the individual rotor thrust $T_i$ and the collective thrust and moment $\mathbf{u}_f$ is given as:
\begin{equation}
\mathbf{u}_f=\mathbf{B}_f\mathbf{f},
\label{eq:uf}
\end{equation}
\begin{equation}
\mathbf{B}_f=\begin{bmatrix}\mathbf{B}_{f,1},\mathbf{B}_{f,2}&\cdots&\mathbf{B}_{f,n_r}\end{bmatrix},
\label{eq:2}
\end{equation}
where the matrix $\mathbf{B}_{f,i}\in\mathbb{R}^{4}$ is the control efficiency matrix of the $i$-th rotor with respect to the CoM of MARS.

\section{Controllability Margin of MARS}
In this section, we define the controllability margin (CM) for MARS~\eqref{eq:system} to quantify its practical controllability. From~\eqref{eq:f} and \eqref{eq:uf}, the constraint set for $\mathbf{u}_f$ can expressed as:
\begin{equation}
\boldsymbol{\varOmega }=\begin{Bmatrix}\mathbf{u}_f\mid\mathbf{u}_f=\mathbf{B}_f\mathbf{f},\mathbf{f}\in U_f\end{Bmatrix}
\label{eq:3}
\end{equation}

For convenience, we define the set of constraints for the control input $\mathbf{u}=\mathbf{u}_f-\mathbf{g}$ as follows:
\begin{equation}
    \mathrm{U=}\left\{\mathbf{u}\mid\mathbf{u}=\mathbf{u}_f-\mathbf{g},\mathbf{u}_f\in\boldsymbol{\varOmega }\right\}
    \label{eq:U_limits}
\end{equation}

We introduce the following controllability for the input-constrained linear systems \eqref{eq:system}, and refer to it as the practical controllability in our paper for convenience. The idea follows from \cite{du2015controllability}.
\begin{lemma}[Practical Controllability~\cite{du2015controllability}]
\label{lem:lemma_Control}
The linear system \eqref{eq:system} under control limits \eqref{eq:U_limits} is (practically) controllable if the following conditions hold:
\begin{enumerate}
    \item $\emph{rank}\,\,\ell\left(\mathbf{A},\mathbf{B}\right)=\!8$, where $\ell\left(\mathbf{A},\mathbf{B}\right)=[\mathbf{B}\!\quad\mathbf{A}\mathbf{B}\!\quad\cdots\quad\mathbf{A}^7\mathbf{B}]$. 
    \item There does not exist a real eigenvector $\mathbf{v}$ of $\mathbf{A}^{\mathrm{T}}$ that satisfies $\mathbf{v}^\mathrm{T}\mathbf{B}\mathbf{u}\leq0$ for all $\mathbf{u}\in\mathrm{U}$. 
\end{enumerate}
\end{lemma}
Condition 1) is the controllability condition for linear unconstrained system. It is easy to verify that Condition 1) of \Cref{lem:lemma_Control} always holds for MARS \eqref{eq:system}. Condition 2) is for all admissible control, depending on the constraint set~\eqref{eq:U_limits}.

Before proceeding, we define the following function that will be used as an indicator to verify~\Cref{lem:lemma_Control},
\begin{equation}
    \hspace*{-0.23cm}
    \left. \mathrm{\zeta}\!\left( \boldsymbol{\alpha },\partial \boldsymbol{\varOmega } \right) \triangleq \left\{ \begin{array}{l}
	\underset{\boldsymbol{\beta }\in \partial \boldsymbol{\varOmega }}{\min} \!\left\| \boldsymbol{\alpha }\!-\!\boldsymbol{\beta } \right\|, \quad \text{if }\boldsymbol{\alpha }\!\in \boldsymbol{\varOmega }\\
	-\underset{\boldsymbol{\beta }\in \partial \boldsymbol{\varOmega }}{\min} \! \left\| \boldsymbol{\alpha }\!-\!\boldsymbol{\beta } \right\|, \quad \text{if } \boldsymbol{\alpha }\!\in \boldsymbol{\varOmega }^{\mathrm{C}} \\
\end{array} \right. \right. 
\label{eq:define CM}
\end{equation}
where $\partial \boldsymbol{\varOmega } $ represents the boundary of the set $\boldsymbol{\varOmega } $, and $\boldsymbol{\varOmega }^C$ represents its complement set. Essentially, it defines a signed distance function from the point $\boldsymbol{\alpha }$ to the surface of $\boldsymbol{\varOmega }$. If the point is inside $\boldsymbol{\varOmega }$, then the distance is non-negative (assuming $\boldsymbol{\varOmega }$ is a closed set without loss of generality); Otherwise, the distance is negative.
% If  $\boldsymbol{\alpha }\in \boldsymbol{\varOmega }^C\cup \partial \boldsymbol{\varOmega } $, then $\zeta (\boldsymbol{\alpha },\partial \boldsymbol{\varOmega })\le 0$ holds. Otherwise, if $\boldsymbol{\alpha }$ is an interior point of $\boldsymbol{\varOmega } $, then $\zeta (\boldsymbol{\alpha },\partial \boldsymbol{\varOmega })\geq 0$. 
Applying~\eqref{eq:define CM} to our problem, if $\mathbf{g}$ is an \textit{interior point} of $\boldsymbol{\varOmega } $, then we have a positive distance: 
\begin{equation}
\zeta \left( \mathbf{g},\partial \boldsymbol{\varOmega } \right) =\min \left\{ \left\| \mathbf{g}-\mathbf{u}_f \right\| ,\mathbf{u}_f\in \partial \boldsymbol{\varOmega } \right\}>0 
\end{equation}

Furthermore, it is easy to show the following equivalent statements.

\begin{lemma}
\label{lem:3equal}For system \eqref{eq:system}, the followings three statements are equivalent:
\begin{enumerate}
    \item There does not exist a real eigenvector $\mathbf{v}$ of $\mathbf{A}^{\mathrm{T}}$ such that $\mathbf{v}^\mathrm{T}\mathbf{B}\mathbf{u}\leq0$ holds for all $\mathbf{u}\in U$ or such that $\mathbf{v}^{\mathrm{T}}\mathbf{B}\big(\mathbf{u}_{f}-\mathbf{g}\big)\leq0$ holds for all $\mathbf{u}_f\in \boldsymbol{\varOmega }$.
    \item $\mathbf{g}$ is an interior point of $\boldsymbol{\varOmega } $.
    \item $\zeta\left(\mathbf{g},\partial\boldsymbol{\varOmega }\right)>0$.
\end{enumerate}
\end{lemma}
According to \Cref{lem:3equal}, Condition 2) of \Cref{lem:lemma_Control} is equivalent to $\zeta\left(\mathbf{g},\partial\boldsymbol{\varOmega }\right)>0$. The latter is much easier to verify.

% \textbf{Theorem 2.2:} A sufficient condition for system \eqref{eq:system} to be controllable is $\zeta\left(\mathbf{g},\partial\boldsymbol{\varOmega }\right)>0$.

It follows that the practical controllability of MARS~\eqref{eq:system} under control constraints \eqref{eq:U_limits} is solely determined by $\zeta(\mathbf{g},\partial\boldsymbol{\varOmega })$. If  $\zeta\left(\mathbf{g},\partial\boldsymbol{\varOmega }\right)<0$, the system is uncontrollable and if $\zeta\left(\mathbf{g},\partial\boldsymbol{\varOmega }\right)>0$, the system is practically controllable under control limits. The larger $\zeta(\mathbf{g},\partial\boldsymbol{\varOmega })$ is, the more control authority the system has. This yields a simple and efficient index to characterize the control performance of MARS.

\section{Robust Self-Reconfiguration of MARS}
In this section, we present a novel post-failure mid-air self-reconfiguration method with three steps for MARS that maximizes its controllability margin (CM), thereby significantly enhancing its fault tolerance capacity. Initially, \Cref{algorithm1} is used to determine the optimal assembly that achieves the highest CM as the objective. Subsequently, we identify the minimum controllable subassembly with faulty units, which ensures that the faulty unit is not isolated during the reconfiguration process. Finally, the reconfiguration step is executed to ensure robustness of the disassembly and assembly following \Cref{algorithm3}. 

For clarity, we denote by $N$ the total number of drone units and by $M$ the number of faulty ones. The structure of MARS is represented by a graph $\mathbf{P}(V,E)$, where the vertices in $V$ correspond to the quadrotors in MARS, and the edges in $E$ represent their pairwise rigid connections. Each vertex is described by the earth coordinates $(x_i, y_i)$ of the corresponding drone unit and its fault conditions. Similarly, $\mathbf{P}^*$  denotes the optimal reconfiguration, and $\mathbf{P}^{\dagger}$ the minimum controllable subassembly. Here, we consider the case where only a single drone failure occurred (denoted by $v_{failure}$, and $M$ is set as 1), and the shape of the structures before and after reconfiguration is the same, which means only the relative locations of drones within the structure can be changed. The algorithm can be potentially generalized to the case of two drone failures.

\vspace{-6mm}
\subsection{Optimal Reconfiguration based on CM}

We assume that the fault status of each rotor is available in real-time, either through direct measurement or soft detection methods. \Cref{algorithm1} shows the procedure to find the optimal target reconfiguration based on the CM index \eqref{eq:define CM}. 
There can be a total number of $A_{N}^{M}$ different reconfigurations. Depending on the shape of the structures, some of these configurations may be symmetric and share the same CM value. These include either axial symmetry along the x- and y-axes (horizontal axes) of the MARS coordinate, or central symmetry around the MARS CoM. Therefore, only one representative reconfiguration is selected from each group of symmetrical configurations for the CM calculation (Line 4-6). For a partial failure unit (e.g. single-rotor failure), notice that the orientation also affects the CM value, thus it will be decided after its position is determined. In Lines 9-15, it checks the other three yaw orientations ($\pi/2$, $\pi$, and $3\pi/2$) for a partial failure quadrotor to further optimize the overall CM for the MARS. 
% \Lin{Update this after revising the Algorithm list 1}.

\begin{algorithm}[t]
    \caption{Find Optimal Reconfiguration}\label{algorithm1}  
\footnotesize
    \renewcommand{\algorithmicrequire}{\textbf{Input:}}
    \renewcommand{\algorithmicensure}{\textbf{Output:}}
    \begin{algorithmic}[1]
        \REQUIRE $\mathbf{P}({V}, {E})$
        \ENSURE $\mathbf{P}^{*}({V}^{*}, {E}^{*})$
            \STATE $N = |{V}|, M=1, \mathbf{C}=\varnothing$
            \FOR{$i=1$ \TO $A_N^M$}
                \STATE $\mathbf{C}_{i}=$  find ${i}^{th}$ possible configuration after failure
                \IF{symmetrical configurations of $\mathbf{C}_{i}$ cannot be found in $\mathbf{C}$}
                    \STATE $\mathbf{C} = \mathbf{C} \cup \mathbf{C}_{i}$
                \ENDIF
            \ENDFOR
            \STATE $\mathbf{P}^{*}({V}^{*}, {E}^{*}) = \mathbf{C}$ with maximum $\mathbf{CM } $ value
            
        \IF{partial failure}
        % \FOR{ $v_{failure}$ in $V_{failure}$}
                \FOR{$\alpha$ = $\pi/2$, $\pi$ and $3\pi/2$}
                \STATE $\mathbf{C}_r = $ new graph after rotating $v_{\text{failure}}$ in $\mathbf{P}^{*}({V}^{*}, {E}^{*})$ at $\alpha$

                \STATE $\mathbf{CM}_{r} = $ Calculate CM ($\mathbf{C}_{r}$)
                \ENDFOR
                \STATE $\mathbf{P}^{*}({V}^{*}, {E}^{*}) = \mathbf{C}_{r}$ with maximum $\mathbf{CM } $ value
            % \ENDFOR
        \ENDIF
        \RETURN $\mathbf{P}^{*}({V}^{*}, {E}^{*})$
    \end{algorithmic} 
\end{algorithm}

\begin{algorithm}[t]
    \caption{Identify the Minimum Controllable Sub-assembly Containing the Faulty Unit}\label{algorithm2}            

\footnotesize
    \renewcommand{\algorithmicrequire}{\textbf{Input:}}
    \renewcommand{\algorithmicensure}{\textbf{Output:}}
    
    \begin{algorithmic}[1]
        \REQUIRE $\mathbf{P}({V}, {E})$, $n=1$, $CM_{max}=-$Inf
        \ENSURE $\mathbf{P}^{\dagger}({V}^{\dagger}, {E}^{\dagger})$
        \WHILE{$CM_{max} < 0$}
            \STATE $\mathbf{C}_{n}=$ the set of sub-assemblies with $n$ functional units in the neighborhood of $v_{failure}$
            \FOR{$\mathbf{C}_{n, i}$ in ${\mathbf{C}_{n}}$}
                \STATE $CM_{max}=\max(CM_{max},$ CM$(\mathbf{C}_{n, i}))$
                \STATE $\mathbf{P}^{\dagger}({V}^{\dagger}, {E}^{\dagger})=\mathbf{C}_{n, i}$ of the maximum $\mathbf{CM } $ value
            \ENDFOR
            \STATE $n=n+1$
            \IF{$n\geq N-M$} 
                \RETURN failure
            \ENDIF
        \ENDWHILE
          
        \RETURN $\mathbf{P}^{\dagger}({V}^{\dagger}, {E}^{\dagger})$
    \end{algorithmic} 
\end{algorithm}

\begin{algorithm}[t]
\footnotesize
\renewcommand{\algorithmicrequire}{\textbf{Input:}}
    \renewcommand{\algorithmicensure}{\textbf{Output:}}
    \caption{Plan Full Disassembly and Assembly Sequence}\label{algorithm3}            
    \begin{algorithmic}[1]
        \REQUIRE $\mathbf{P}({V}, {E}), \mathbf{P}^{*}({V}^{*}, {E}^{*}), \mathbf{P}^{\dagger}({V}^{\dagger}, {E}^{\dagger})$
        \ENSURE $\mathbf{P}({V}, {E})$ 
        \STATE $V_{disassembly}= V\setminus V^{\dagger}$

        \WHILE{$|V_{disassembly}|>0$}
            \FOR{$v_i$ in $V_{disassembly}$} 
            % disassembling stage
                \STATE $\mathbf{P}_{temp} = \mathbf{P}$ after deleting $v_i$
                \STATE set $v_{target}$ = $v_i$ s.t. $\mathbf{P}_{temp}$ has maximum CM
            \ENDFOR
            
            \STATE delete $v_{target}$ from $\mathbf{P}$ and $V_{disassembly}$
            
            % assembling stage
            \FOR{$v^*_i$ in $V^*$} 
                \IF{$v^*_i$ is a legal transfer target}
                    \STATE $\mathbf{P}_{temp}=$ adding $v_i$ to the position of $v^*_i$ in $\mathbf{P}$
                    \STATE set $v^*$ = $v^*_i$ s.t. $\mathbf{P}_{temp}$ has maximum CM
                \ENDIF
            \ENDFOR
            \STATE add $v_{target}$ to the position of $v^*$ in $\mathbf{P}$ 
        \IF{$\mathbf{P}$ has negative CM after detaching and docking}
            \RETURN failure
        \ENDIF
        \ENDWHILE
    \RETURN $\mathbf{P}$ 
\end{algorithmic} 
\end{algorithm}

\subsection{Disassembly and Assembly Sequence Planning}
With the target reconfiguration obtained in~\Cref{algorithm1}, we propose the planning algorithm in~\Cref{algorithm2} and~\Cref{algorithm3}  for the reconfiguration sequence. During the process, there can be multiple choices; Priority is given to steps that result in a smaller decrease in the assembly’s CM value when disassembling and a greater increase when assembling. The specific steps for self-reconfiguration are as follows.

% \subsubsection{Generate the Minimal Controllable Sub-assembly Containing the Faulty Units to be Transferred (\Cref{algorithm2})}
% This step aims to ensure that the faulty unit is controllable by establishing connections with other functional units, rather than being handled in isolation. To facilitate a flexible process, the smallest controllable subassembly with positive CM value containing the faulty unit is identified. If the original configuration is the minimal controllable subassembly (see lines 8-10), it cannot transform into the optimal reconfiguration without landing, as any detachment would make the rest assembly uncontrollable. 
% Taking a 3$\times$2 assembly composed of six units as an example (See top left of Fig.~\ref{fig:Self-reconfiguration flow} for the indices of the drone units in the original assembly). For case 1, the complete faulty unit only connects to one normal unit No.4. This assembly is uncontrollable because s the torque generated by the
% gravity and thrust cannot be compensate. In case 2, we avoid the situation by creating minimum controllable sub-assembly containing the faulty unit to transferring the faulty unit.

% This step aims to ensure that the faulty unit remains controllable by establishing connections with other functional units rather than handling it in isolation. 

\subsubsection{Generating the Minimal Controllable Subassembly Containing the Faulty Units for Transfer (\Cref{algorithm2})}
\begin{figure}[!t]
\centering
\includegraphics[width=3in]{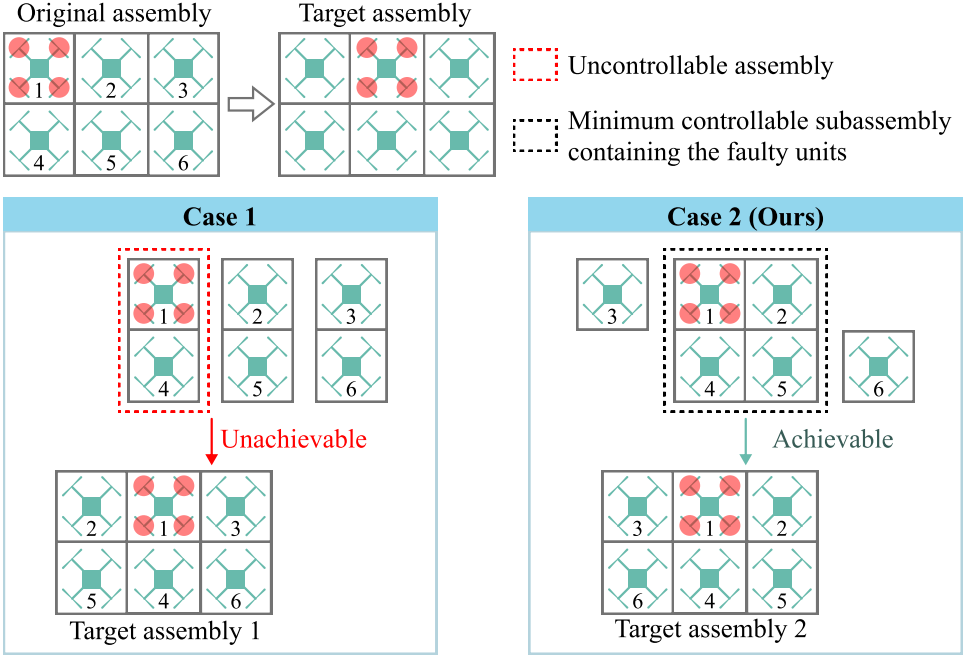}
\caption{Self-reconfiguration flow after the failure of drone unit 1 in a 3$\times$2 assembly. In Case 1, simply connecting a normal unit adjacent to the complete faulty unit is not controllable, as the torque generated by the gravity and thrust cannot be compensated, resulting in negative CM. In Case 2, the minimum controllable sub-assembly containing the faulty unit is identified to assist in transferring the faulty unit.}
\label{fig:Self-reconfiguration flow}
\vspace{-3mm}
\end{figure}

To enable a flexible reconfiguration process, the smallest controllable subassembly (with a positive CM value) that includes the faulty unit is identified. If the original configuration already forms the minimal controllable subassembly (see lines 8-10), it cannot be reconfigured optimally without landing, as any detachment would render the remaining assembly uncontrollable. Consider a $3\times2$ assembly composed of six units (see the top left of Fig.~\ref{fig:Self-reconfiguration flow} for the indices of the drone units in the original assembly). In Case 1, the fully faulty unit is connected to only one normal unit, No.~4. This assembly is uncontrollable because the torque generated by gravity and thrust cannot be compensated. In Case 2, this issue is avoided by forming the minimal controllable subassembly containing the faulty unit before transferring it.

\subsubsection{Full Self-Reconfiguration Sequence Planning (\Cref{algorithm3} )}
To decide the optimal disassembly and assembly sequence, \Cref{algorithm3} first finds all units in $\mathbf{P}$ that need to be detached (Line 1). We always disassemble $v_{target}$ that causes the smallest CM reduction of the remaining structure, and select the target position $v^*$  if the CM increase of the generated assembly is the largest (see lines 4-12). These steps will continue until the optimal assembly is achieved. For structures $m \times n$, we can transfer more than one unit at a time, which is called the partial disassembly approach in contrast to the full disassembly approach, where normal units are transferred one at a time. Practically, partial disassembly can be implemented by iteratively finding substructures that appear in the original and optimal configurations.

\section{Evaluation}

We employ a high-fidelity quadrotor model in CoppeliaSim~\cite{coppeliaSim} for simulation studies and evaluations. Tests are conducted on the controllability of different assemblies, the CM during unit failures, and the control performance after self-reconfiguration. Moreover, we mainly consider and simulate complete failure cases rather than rotor failures, since it is more difficult to implement reconfiguration and fault-tolerant control, and often requires more precise controllability analysis.

\begin{figure}[!t]
\centering
\includegraphics[width=3.4in]{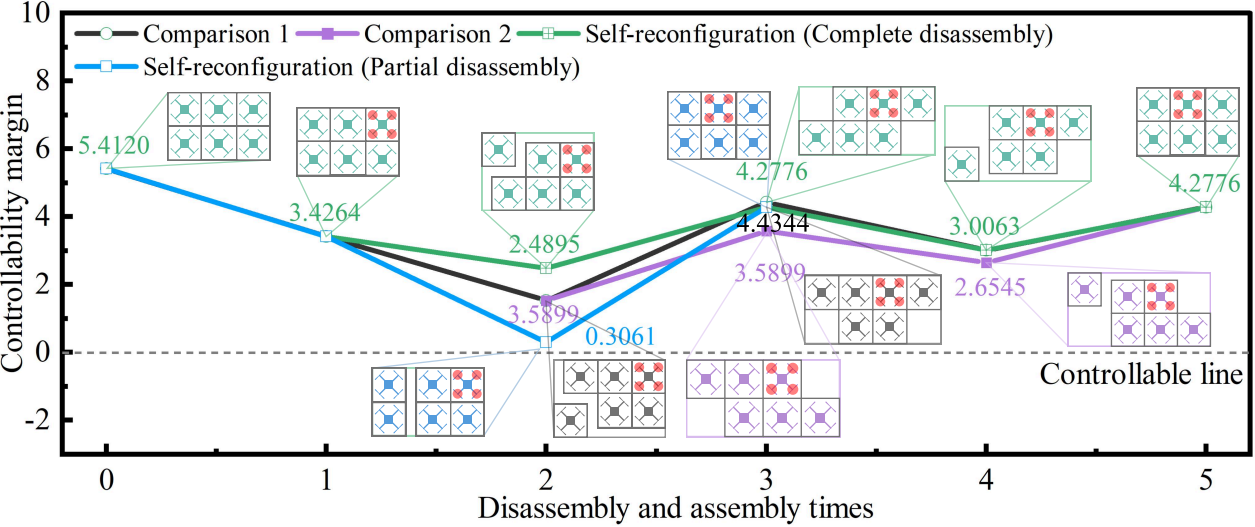}
\caption{Self-reconfiguration flow after failure of unit No.3 in a 3$\times$2 assembly. Black and purple lines show different disassembly and reassembly choices, and the green line indicating the highest CM case from our algorithm. Full disassembly moves and reconfigures units one by one, while partial disassembly uses sub-assemblies.}
\label{fig:Self-reconfiguration flow after failure of unit 1 in a 3x2 assembly}
\vspace{-3 mm}
\end{figure}

\begin{figure}[!t]
\centering
\includegraphics[width=3.4in]{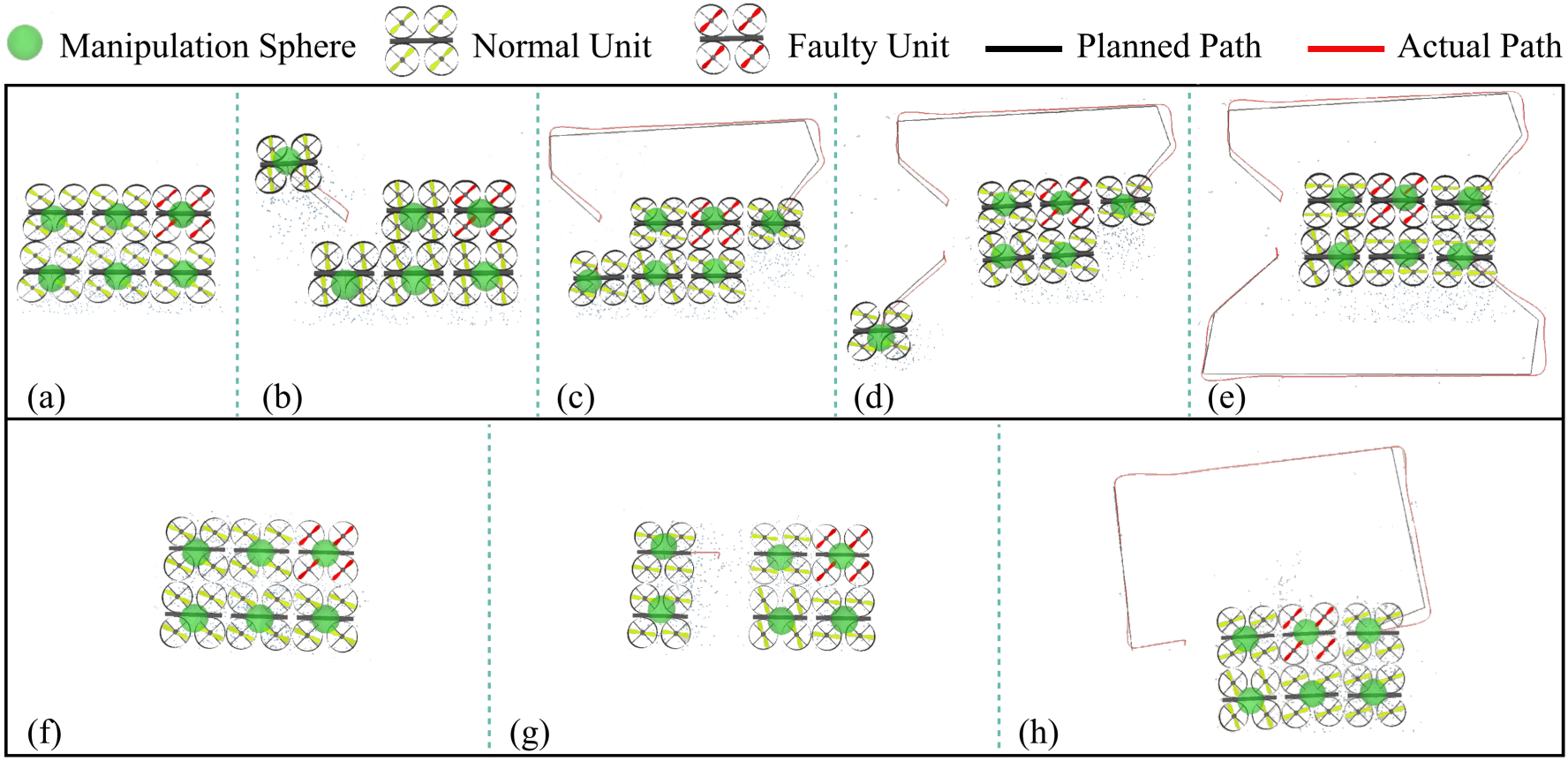}
\caption{Dynamical simulation of self-reconfiguration processes in CoppeliaSim. First row: (a)--(e), full disassembly approach. Second row: (f)--(h), partial disassembly approach. Additional demonstrations, including rotor failures and alternative configurations, are available at \href{https://github.com/RuiHuangNUS/MARS-Reconfig}{https://github.com/RuiHuangNUS/MARS-Reconfig}. }
\label{fig:Simulation snapshots of self-reconfiguration}
\vspace{-6mm}
\end{figure}

\begin{figure}[!t]
\centering
\includegraphics[width=3.4in]{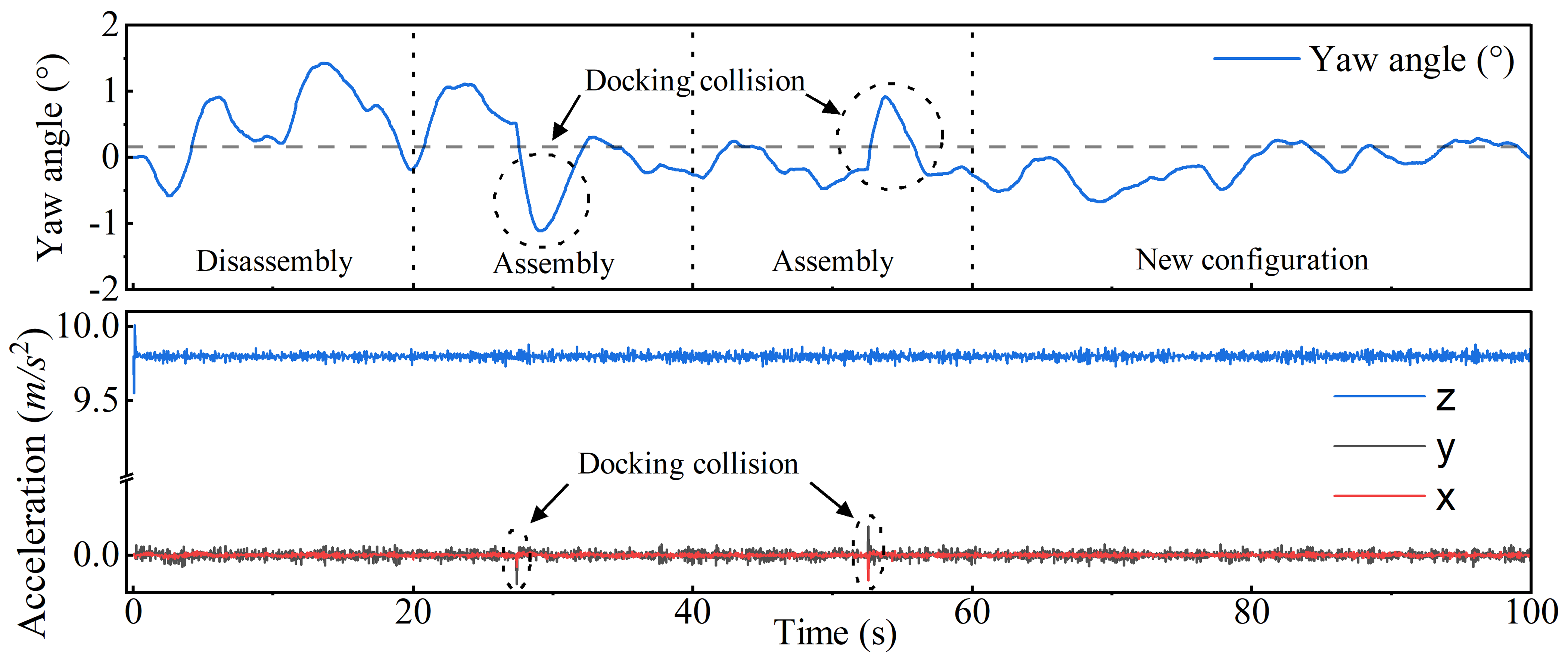}
\caption{Yaw angle and IMU-measured accelerations of the faulty unit during the self-reconfiguration (full disassembly approach, corresponding to (a)-(e) in Fig.~\ref{fig:Simulation snapshots of self-reconfiguration}).}
\label{fig:Angle and acceleration of the faulty unit}
\vspace{-2mm}
\end{figure}
\begin{figure}[!t]
\centering
\includegraphics[width=3.4in]{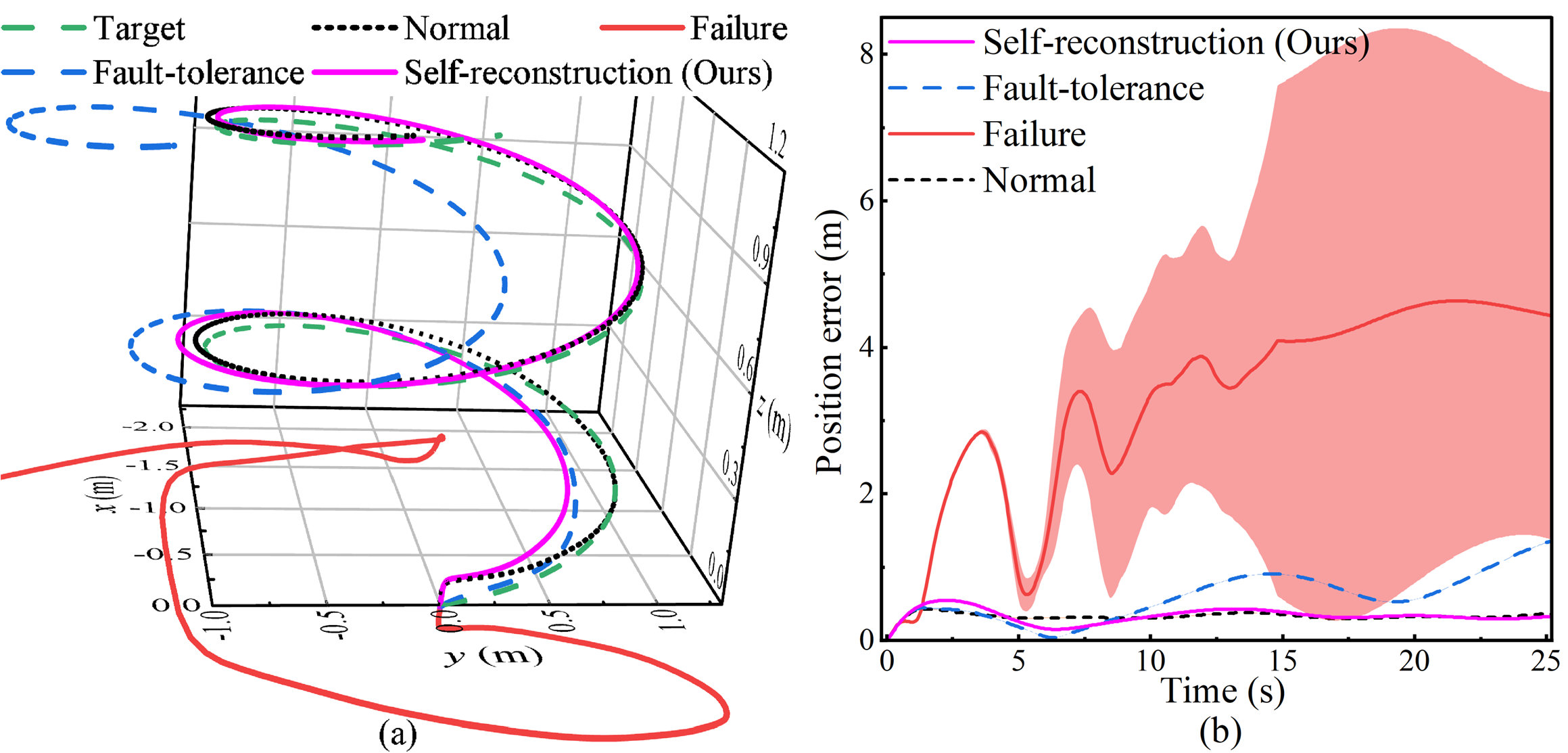}%
\hfil

\caption{The comparison of trajectory tracking before and after self-reconstruction following a complete failure. (a) Trajectory tracking. (b) Position error.}
\label{fig: 1 unit complete failure}
\vspace{-2mm}
\end{figure}

\begin{figure}[!t]
\centering
\includegraphics[width=3.2in]{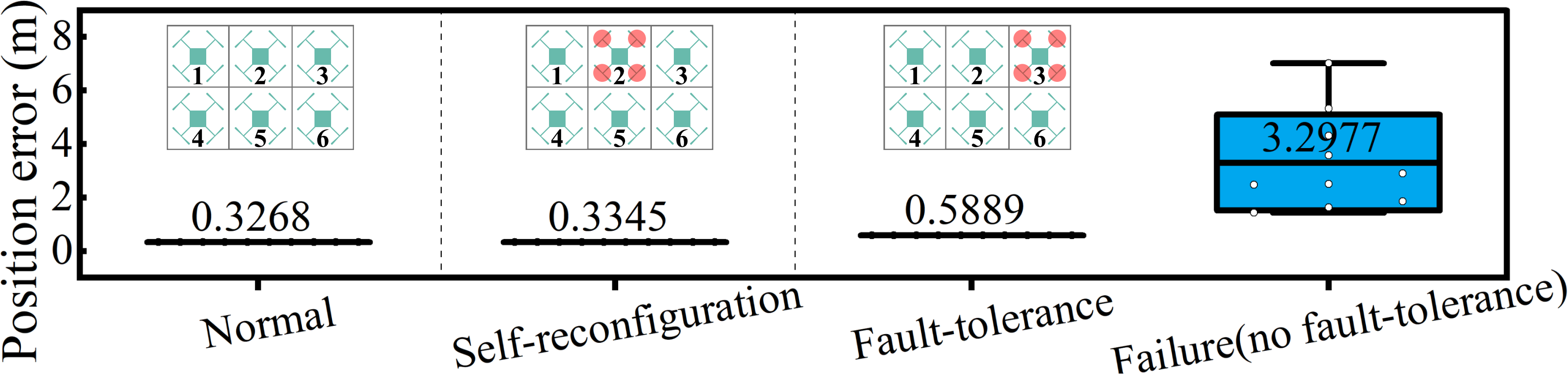}
\caption{Comparison of trajectory tracking error statistics under different conditions: (1) fully functional (Normal), (2) self-reconfigured assembly with fault-tolerant control (FTC), (3) original faulty assembly with FTC, and (4) original faulty assembly without FTC. The plots for (1)--(4) are arranged from left to right.}
\label{fig:CM and Performance indicator}
\vspace{-6mm}
\end{figure}
\subsection{Robust Self-Reconfiguration}

Severe issues such as communication breakdowns, battery depletion, or controller malfunctions can result in the complete failure of a drone unit. In this subsection, we analyze the CM of the assembly and the impact on its trajectory tracking before and after self-reconfiguration in the event of complete failures. 

\subsubsection{Complete Failure of One and Two Units}

\Cref{table:failure 1 or 2 unit} shows the CM of the assembly with faulty units in different locations, where $n_{i}=0$ means the $i$-th drone is completely faulty. Symmetrical failures yield identical CM values and therefore are collected in one row. With any single faulty drone unit, the assembly is still controllable (See Rows 2 and 3). Note that corner unit failures (Row 3) can generate larger failure moments than central units (Row 2), resulting in lower CM values. The table also listed CM values for two faulty units (See Rows 4--8).
% This validates the proposed CM analysis in characterizing the practical controllability and the measure of available control authority.

Fig.~\ref{fig:Self-reconfiguration flow after failure of unit 1 in a 3x2 assembly} compares the intermediate stage CM values of different self-reconfiguration sequences, transferring the faulty Unit No. 3 at the top right corner in a 3$\times$2 assembly to its middle location at Unit No. 2. The proposed method (green line) maximizes CM at each intermediate stages. This approach improves CM by 8.88\% compared to the black line sequence and 25.62\% compared to the purple line sequence, enhancing safety and reliability. The partial self-reconfiguration method further reduces operations by 50\% compared to full self-reconfiguration. 
%\Lin{what methods are used by black line and purple line? what do you mean by partial and complete reconfiguration?}

\begin{table}[!t]
\caption{Controllability analysis: failures in a 3$\times$2 assembly.\label{table:failure 1 or 2 unit}}
\resizebox{\columnwidth}{!}{
\centering
\begin{tabular}{c|c|c c}
\hline
No. &\textbf{Unit Failure}& \textbf{CM}& \textbf{Controllability}\\\hline
  1&No failure& 5.4120
& Controllable\\\hline
 2&$n_{i}=0, i\in\{2,5\}$
& 4.2776
& Controllable\\\hline
 3&$n_{i}=0, i\in\{1,3,4,6\}$
 & 3.4264
& Controllable\\\hline
  4&\begin{tabular}[c]{@{}c@{}}
 Failure Drones: (1 and 6), (3 and 4),\\ or (2 and 5)
 \end{tabular}& 2.7473&Controllable\\\hline
  5&\begin{tabular}[c]{@{}c@{}}
 Failure Drones: (1 and 5), (3 and 5),\\  (2 and 4), or (2 and 6)
 \end{tabular}& 2.4106&Controllable\\\hline
  6&\begin{tabular}[c]{@{}c@{}}
 Failure Drones: (1 and 2), (5 and 6), \\ (2 and 3), or (4 and 5)
 \end{tabular}
 & -0.0037 &Uncontrollable   
\\\hline
   7&Failure Drones: (1 and 4) or (3 and 6)              & -0.0286 &Uncontrollable   
\\\hline
  8&Failure Drones: (1 and 3) or (4 and 6)           & -0.0676 &Uncontrollable   \\\hline
\end{tabular}
}
\vspace{-5mm}
\end{table}

\begin{figure*}[!t]
\centering
\includegraphics[width=6.9in]{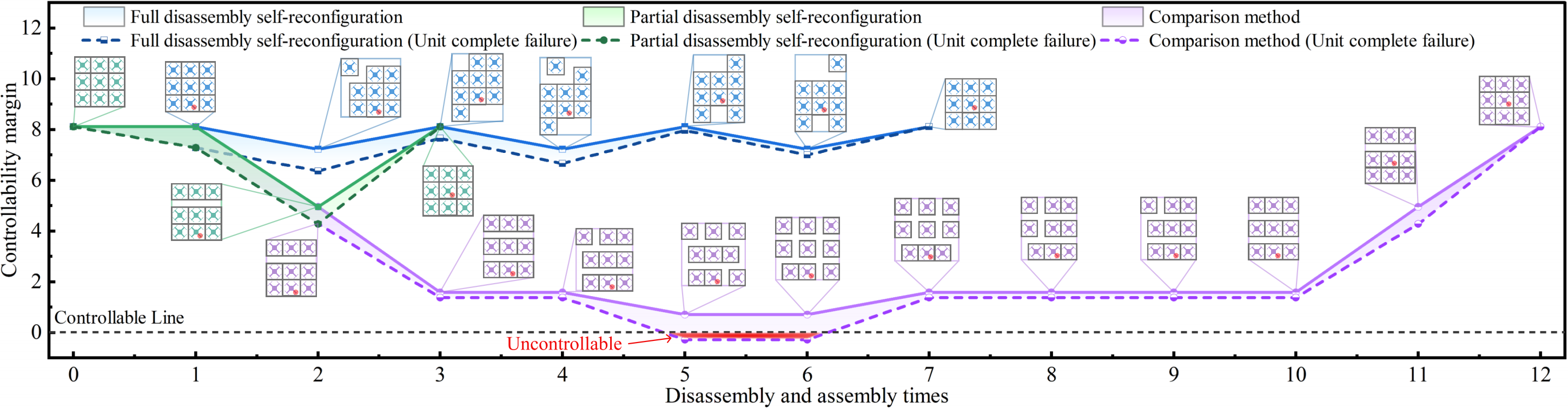}
\vspace{-3mm}
\caption{Comparison of disassembly and assembly times and CM for failure at position 8 in a 3$\times$3 assembly. The full disassembly self-reconfiguration (green line) indicates the highest CM case, and partial disassembly method (blue line) has relatively lower CM value with fewer steps. In contrast, when complete failure occurs, the method indicated by the purple dotted line can be uncontrollable with negative CM.}
\label{fig:Comparison in a 3x3 assembly}
\vspace{-4mm}
\end{figure*}

As in Fig.~\ref{fig:Simulation snapshots of self-reconfiguration}, we also implement the dynamical simulation of full (first row, (a)--(e)) and partial (second row, (f)--(h)) disassembly approaches. As no simulation results are available for Algorithm 2 in \cite{gandhi2020self}, and the code is not open-source, the comparison is primarily conducted against their Algorithm 1. Unlike the previous method, which fully disassembles the structure first then do reassembly, our full disassembly directly reassembles units into the partially faulty sub-assembly immediately after disassembly, resulting in higher CM and fewer reconfiguration steps as in Fig. ~\ref{fig:Comparison in a 3x3 assembly}. We also plot yaw angle and IMU-measured accelerations of the full disassembly approach in Fig.~\ref{fig:Angle and acceleration of the faulty unit}. Due to the impact of rigid connections, approximately 1° yaw deviations occur during docking of Units No.1 and No.4, indicating units are stable during docking process. The acceleration spikes caused by the impact can also be observed in the $x$ and $y$ axes. The self-reconfiguration of an assembly in response to rotor failure is discussed in \cite{huang2025robust}.

Fig.~\ref{fig: 1 unit complete failure} shows the trajectory tracking performance of a spiral curve under different settings. Note that the continuous control strategy must be adjusted after a faulty unit is detected. We adopted the fault-tolerant control (FTC) strategy in our previous work to redistribute the collective control command $\mathbf{u}_{f}$ into individual functional unit's control command (total thrust and torques). In the simulated case, drone Unit 3 completely fails, generating off-center moments that disrupt the flight. The purple solid curve represents the trajectory under self-reconfiguration with FTC, demonstrating a tracking performance that closely matches that of the fully functional assembly (black dot-dash curve with legend Normal). The blue dot-dash curve denotes the trajectory under the original configuration with FTC (with legend Fault-tolerance). It can be seen that self-reconfiguration with FTC significantly improves the trajectory tracking compared to the original configuration with FTC. \textbf{This demonstrates the effectiveness of the proposed method in enhancing the robustness of tracking control}. In addition, the red curve (with the legend Failure) represents the trajectory under the original configuration but without the fault-tolerant control. The MARS immediately deviates from the target after takeoff and is out of control. In addition, we plot in Fig.~\ref{fig: 1 unit complete failure}(b) the average (solid curve) and the distribution (shaded areas) position errors (in RMSE) from 10 repeated experiments in different settings. Fig.~\ref{fig:CM and Performance indicator} shows the box plot of the errors.

% Fig.~\ref{fig:CM and Performance indicator} 
%  summarizes performance metrics from 10 trajectory tracking experiments under complete unit failure. Self-reconfiguration reduced average position error, with fault-tolerant control decreasing error by 82.14\% (from 3.2978 m to 0.5889 m), and self-reconfiguration further reduced it by 43.19\% to 0.3345m. The error increase compared to the no-failure condition (0.3268 m) was only 2.36\%. These results highlight improved flight performance with self-reconfiguration and its potential to enhance safety and reliability in multi-unit assemblies. The strong correlation between CM and tracking error suggests that the CM metric can guide optimal post-failure self-reconfiguration.

\subsubsection{Rotor Failure and Performance Degradation}
In previously analyzed scenarios with complete failure units, our approaches suggest that relocating the failed units to CoM generally maximizes the controllability. However, the position of faulty units with partial failures can have a significant impact on CM, where rotors can still generate lift forces but with a degraded efficiency. For example, considering a 3$\times$2 assembly, we denote by $\eta_{i,j} \in[0,1]$ the efficiency coefficient of the $j$-th rotor in the $i$-th unit, where $\eta_{i,j}=1$ means a fully efficient rotor without fault and $\eta_{i,j}=0$ means complete failure. 
We assume drone Unit 4 suffers partial failures. As shown in Table~\ref{table:Thruster failure 3x2}, changes in the position of the faulty rotors in Unit 4 result in CM values ranging from 4.9504 to 5.2697, highlighting the necessity of the rotating in Algorithm~\ref{algorithm1}. 
% \Lin{what can this prove? I suppose you should give examples where the partially faulty unit should be placed in the corner instead of in the center. To fulfill this purpose, we should compare the CM values of these two different cases}
% complicating the aforementioned empirical conclusions. When multiple partially faulty units are present, the empirical conclusion of placing the defective unit at the center no longer holds. Complex assemblies with non-standard shapes, customized for payloads, further complicate this. 
\begin{table}
\centering
\caption{Controllability Analysis: Rotor Failures in 3$\times$2 Assembly.\label{table:Thruster failure 3x2}}
\begin{tabular}{c|c c} 
\hline
\textbf{Rotor Partial Failure}& \textbf{CM}     & \textbf{Controllability}  \\ 
\hline
\begin{tabular}[c]{@{}c@{}}$\eta_{1,1}=0$\end{tabular}& 5.2697& Controllable     \\ 
\hline
\begin{tabular}[c]{@{}c@{}}$\eta_{1,2}=0$\end{tabular}& 4.9604& Controllable     \\ 
\hline
\begin{tabular}[c]{@{}c@{}}$\eta_{1,3}=0$\end{tabular}& 4.9504& Controllable     \\ 
\hline
\begin{tabular}[c]{@{}c@{}}$\eta_{1,4}=0$\end{tabular}& 4.9604& Controllable     \\\hline
\end{tabular}
\vspace{-5mm}
\end{table}

%\textbf{Remark 1:} The performance degradation or partial failure of a unit's thruster is more easily addressed by the self-reconfiguration method proposed in this paper compared to a complete failure. Moreover, the self-reconfiguration calculation process and fault-tolerant control are consistent with those for a complete failure, yielding similar results. Therefore, this paper does not further discuss the algorithm's performance in the case of a thruster failure within a unit.  

\subsection{Comparison with the baseline method}
\subsubsection{Practical Controllability}
To further demonstrate the robustness of our method, we compare the CM values at the intermediate stages of different approaches in a 3$\times$3 self-reconfiguration scenario with one faulty unit. In Fig.~\ref{fig:Comparison in a 3x3 assembly}, the self-reconfiguration sequences generated by different approaches are plotted, along with the line charts representing their intermediate CM values. Solid lines present scenarios with a single rotor failure, while dashed lines denote complete unit failure cases. Our proposed methods consistently achieve higher CM, with improvements of up to 264.50\% (blue) and 138.63\% (green) compared to~\cite{gandhi2020self}. In contrast, the method in~\cite{gandhi2020self}, which does not account for practical controllability, plans uncontrollable subassemblies in complete unit failure cases, resulting in a CM of $-0.2824$ (see the red line between assembly times 5 and 6). These uncontrollable subassemblies are practically uncontrollable and lead directly to a crash in reality. In contrast, the proposed robust methods guarantee practical controllability at each intermediate step, ensuring successful self-reconfiguration.

\subsubsection{Disassembly and Assembly Times}

 As seen at the end of each line in~Fig.~\ref{fig:Comparison in a 3x3 assembly}, both our methods and the comparison method from~\cite{gandhi2020self} identify the same optimal target assembly (at Steps 3, 7, and 12, respectively). However, our methods generally require significantly fewer steps. The method in~\cite{gandhi2020self} involves eleven steps (2-12) for reconfiguration after a fault occurs, while our proposed methods reduce the total steps to two (Steps 2 to 3 for the partial disassembly approach) and six (Steps 2 to 7 for the full disassembly approach), cutting the number of steps by 81.81\% and 45.45\%, respectively. This improved efficiency enhances system safety by minimizing altitude and attitude disturbances during the assembly and disassembly and also optimizes battery usage~\cite{saldana2018modquad,carlson2022multi,saldana2019design}.
 
\subsubsection{Time Complexity}
Considering an $m \times n$ rectangular structure with one unit failure, at most $mn$ iterations are required to compute the faulty unit's optimal position in the reconfiguration~\Cref{algorithm1}. For~\Cref{algorithm2}, the time complexity is $O(m^2n^2)$, which corresponds to the total number of all subassemblies~\cite{gandhi2020self}. For~\Cref{algorithm3}, at iteration $k$, there are $mn-k$ units left at the worst case. It takes $O(mn)$ to determine the next disassembly unit and its optimal position in the reconfiguration. Hence, summing over $k$, the overall complexity is $O(m^{2}n^2)$. Since all three algorithms execute sequentially, the overall time complexity remains $O(m^2n^2)$. This provides a more efficient reconfiguration compared to the previous full disassembly method, which has a complexity of $O((m+n)m^2n^2)$ \cite{gandhi2020self}.

\section{Conclusion}
To mitigate rotor and unit failures in modular aerial robotic systems (MARS), this paper presents a robust self-reconfiguration method to further enhance fault tolerance control. Our approach is based on maximizing the controllability margin of the MARS under constrained control limits. We developed algorithms that first identify the minimum controllable subassemblies containing the fault units and then plan the disassembly and reassembly sequence with maximum controllability at each stage. Our approach demonstrated strong robustness with a high degree of controllability in simulations compared to the existing result and significantly reduced trajectory tracking error with the re-configured assembly.

\bibliographystyle{IEEEtran}
% Generated by IEEEtran.bst, version: 1.14 (2015/08/26)

% \bibliography{reference}

%\begin{thebibliography}{99}

% \end{thebibliography}

\end{document}